\documentclass[10pt,twocolumn,letterpaper]{article}

\usepackage[pagenumbers]{cvpr} 

\usepackage{graphicx}
\usepackage{amsmath}
\usepackage{amssymb}
\usepackage{booktabs}
\usepackage{makecell}
\usepackage{multirow}

\usepackage[pagebackref,breaklinks,colorlinks]{hyperref}

\usepackage[capitalize]{cleveref}
\crefname{section}{Sec.}{Secs.}
\Crefname{section}{Section}{Sections}
\Crefname{table}{Table}{Tables}
\crefname{table}{Tab.}{Tabs.}

\def\cvprPaperID{1985} 
\def\confName{CVPR}
\def\confYear{2023}

\usepackage{cuted}
\usepackage{capt-of}

\begin{document}

\title{TextIR: A Simple Framework for Text-based Editable Image Restoration}

\author{%
  Yunpeng Bai$^{1}$,  Cairong Wang$^{1}$, Shuzhao Xie$^{1}$, Chao Dong$^{2,3}$, Chun Yuan$^{1,4}$, Zhi Wang$^{1,4}$ \\[0.5em]
  $^{1}$ Tsinghua University, $^{2}$Shenzhen Institutes of Advanced Technology, Chinese Academy of Sciences\\ $^{3}$ Shanghai AI Laboratory, China, $^{4}$Peng Cheng Laboratory, Shenzhen, China \\[0.3em]
}

\maketitle

\begin{strip}
\centering
\includegraphics[width=\textwidth]{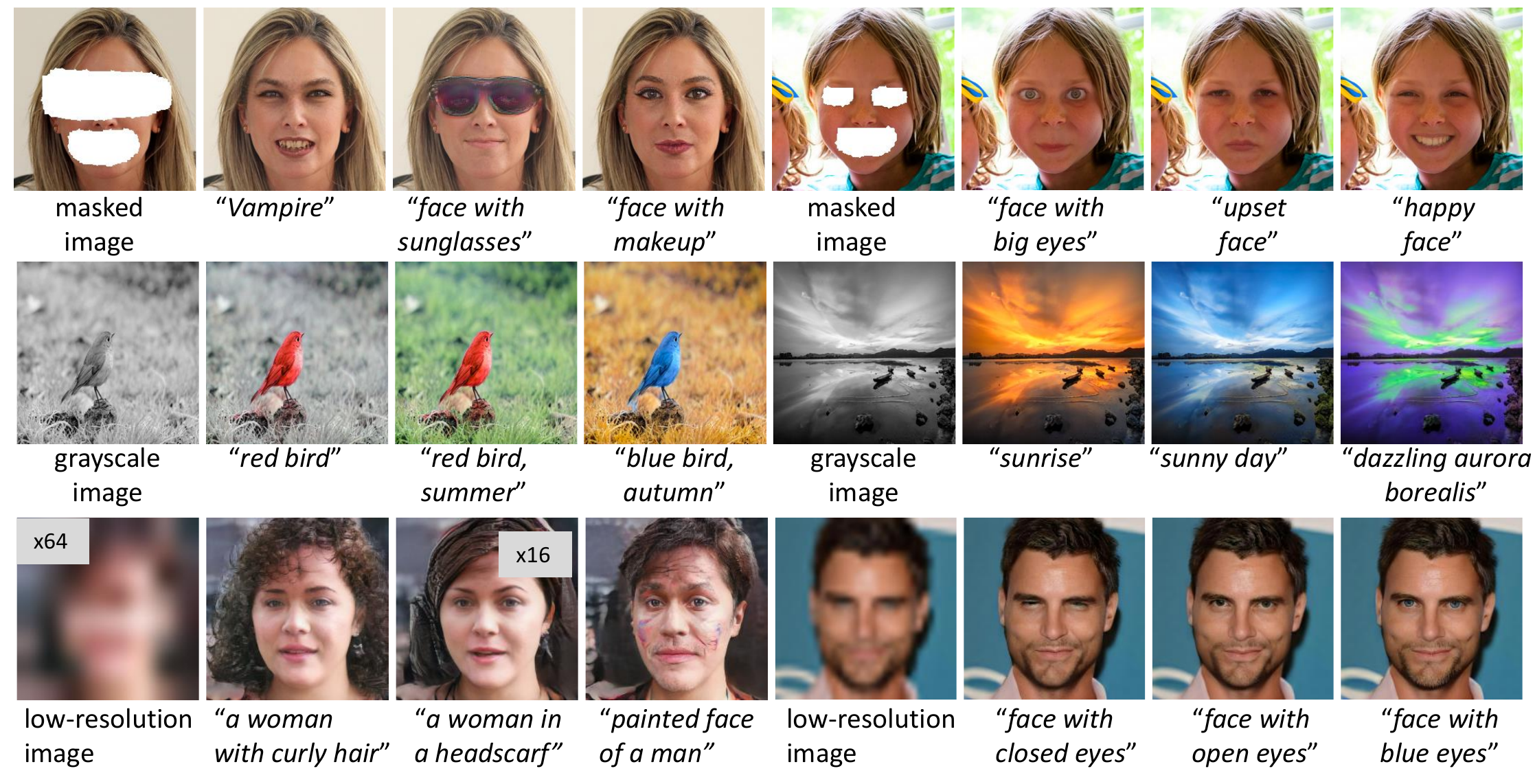}
\captionof{figure}{\textbf{Overview of image restoration results of the proposed framework.} The framework can be applied to various restoration tasks. For a degraded image, users can use different text inputs to get the restored image they want. Our framework can accept a wide range of text input and can precisely locate the area to be restored.}
\label{fig:teaser}
\end{strip}


\begin{abstract}
Most existing image restoration methods use neural networks to learn strong image-level priors from huge data to estimate the lost information. However, these works still struggle in cases when images have severe information deficits. Introducing external priors or using reference images to provide information also have limitations in the application domain. In contrast, text input is more readily available and provides information with higher flexibility. In this work, we design an effective framework that allows the user to control the restoration process of degraded images with text descriptions. We use the text-image feature compatibility of the CLIP to alleviate the difficulty of fusing text and image features. Our framework can be used for various image restoration tasks, including image inpainting, image super-resolution, and image colorization. Extensive experiments demonstrate the effectiveness of our method. This new framework also provide a good starting point for the text-based image restoration task.

\end{abstract}

\section{Introduction}
\label{sec:intro}

Image restoration is a fundamental computer vision problem, which takes a degraded image (e.g., grayscale, damaged, low-resolution image) as input and reconstructs the corresponding high-quality image. Due to its ill-posed nature, most existing works propose complex deep models to learn strong image priors from massive data to fill in the lost information. 
However, both convolutional neural networks (CNNs) and Transformer architectures struggle to deal with cases that contain severe information deficits. For example, when the input contains large holes or has complex content (e.g., semantic layout, texture, and depth) to be restored, these methods cannot generate visual-pleasant images.


To further improve the performance of extreme cases, some external priors are explored and introduced into the image restoration model to provide additional guidance. These priors usually include generative prior \cite{yang2021gan, wang2021towards} and structural prior \cite{chen2018fsrnet, buhler2020deepsee}, but they can only be applied to a particular domain, such as face, and cannot be generalized to other image contents. 
Some other works have attempted to solve these hard cases by using another reference image with some desired content that is useful to restore the degraded image. However, they still require users to find a suitable reference image first, which will limit its application scenarios.



Compared to images, text descriptions can easily represent the concept of the image that matches our imagination.
It can effortlessly depict an image's global style, local property, and abstract concepts, such as color, shape, emotion, etc. 
Therefore, we can use text descriptions to provide the information needed for the restoration process more flexibly, improve the controllability of image restoration methods, and achieve editable restoration effects to meet diverse requirements. Some approaches attempt to use text to inpaint \cite{zhang2020text,lin2020mmfl} or colorize \cite{weng2022code} images. However, these methods have strict requirements for the dataset, such as images with text that accurately describe their colors. Additionally, their models can only be applied to a specific task.

How to effectively fuse the features of the two data modalities of text and image is another challenge.
The Contrastive Language-Image Pre-training (CLIP) model \cite{radford2021learning} is a recent advance that connects the image and text data by training two encoders on an Internet-scale dataset. CLIP learns a multi-modal embedding space shared by text and image
features and contains a wide range of visual concepts.
In this work, we utilize the properties of CLIP and propose the first text-based image restoration framework — TextIR. Specifically, TextIR uses the text-image feature compatibility of CLIP to convert text descriptions into the corresponding image features, which are then merged with the degraded image features for restoration. It is worth noting that the training procedure does not require additional text-image pairs. This framework allows users to control the restoration process with text descriptions and can be used for various image restoration tasks, including image inpainting, image super-resolution, image colorization, etc.

To summarize, our main contributions are as follows:
\begin{itemize}
\item We design a simple and effective framework that allows the user to use text input to get desired image restoration results.

\item We take advantage of CLIP's text-image feature compatibility to enable a better fusion of image and text features.

\item Our framework can be used for various image restoration tasks, including image inpainting, image super-resolution, and image colorization.
\end{itemize}

\section{Related Works}
\label{sec:related}
\subsection{Image Restoration}
Image restoration aims to remove the effects of degradation from the degraded image input and reconstructs the original high-quality image . In recent years, CNN-based architectures \cite{ zamir2020learning,  dudhane2022burst}, along with spatial \& channel attention modules \cite{gu2019self, liu2019dual, zamir2020learning, zhang2019residual} and skip connection-based approaches \cite{li2018recurrent, zamir2020learning}, have achieve significant breakthrough in this task. In addition, encoder-decoder based U-Net architectures \cite{abuolaim2020defocus, cho2021rethinking} have been predominantly studied for restoration due to their hierarchical multi-scale representation while remaining computationally efficient. So far, replacing the CNN with Transformers \cite{vaswani2017attention} that have the capacity to capture long-range dependencies in the data enables researchers \cite{liang2021swinir, zamir2022restormer, wang2022uformer} to achieve better performance. However, these methods still perform poorly on severely degraded images. 


To tackle this drawback, subsequent works intend to introduce additional prior as guidance. As a pioneer in the use of generative priors, GFP-GAN \cite{wang2021towards} utilizes rich and diverse priors encoded in a pre-trained StyleGAN for blind face restoration. Thereafter, GPEN \cite{yang2021gan} trains a GAN for generating high-quality face images and embeds it into a U-shaped network as the prior decoder. Contemporary works integrate face structure priors \cite{ chen2018fsrnet, buhler2020deepsee} into restoration. For instance, FSRNet \cite{chen2018fsrnet} uses a prior estimation network to ensure that the spatial information at different scales is preserved in the process of face super-resolution. DeepSEE \cite{buhler2020deepsee} explores the application of semantic maps in the face super-resolution method. Nevertheless, these prior-based methods are only applicable to images of a particular domain, which limits the application scenarios. Our framework uses text descriptions to provide the information needed for the restoration process more flexibly and can be used on a variety of data categories.


\subsection{Text-driven Image Manipulation}
With the successful development of cross-modal visual and linguistic representations \cite{lu2019vilbert,su2019vl,tan2019lxmert,yuan2021florence}, especially the omnipotent CLIP \cite{radford2021learning}, many efforts \cite{chen2018language, jiang2021talk, patashnik2021styleclip, wang2022clip, kwon2022clipstyler, xia2021tedigan, wei2022hairclip} have recently started investigating text-driven image manipulation. However, there are no existing methods specifically for image restoration. Among these works, the most relevant ones are StyleCLIP \cite{patashnik2021styleclip}, HairCLIP \cite{wei2022hairclip}, and CLIPStyler \cite{kwon2022clipstyler}. StyleCLIP performs attribute manipulation with exploring learned latent space of StyleGANv2 \cite{karras2020analyzing}. However, it can only edit the original image and is limited in the specific domain. Therefore, CLIPStyler proposes a framework that enables a style transfer only with a text description of the desired style. Besides, HairCLIP introduces a method tailored for hair editing, which can manipulate hair attributes individually or jointly based on the texts or reference images. These methods can only edit images according to the attributes in the text description, and cannot use the information in the text to restore the degraded image.

\subsection{Text-based Image Generation}
In recent years, there has been a rapid rise in text-based image generation works. Early RNN-based works \cite{mansimov2015generating} were quickly superseded by generative adversarial approaches \cite{reed2016generative}. The latter was further improved by multi-stage architectures \cite{zhang2017stackgan, zhang2018stackgan++} and an attention mechanism \cite{xu2018attngan}. DALL-E \cite{ramesh2021zero, ramesh2022hierarchical} introduced a GAN-free two-stage approach. First, a discrete VAE \cite{razavi2019generating, van2017neural} is trained to reduce the context for the transformer. Next, a Transformer \cite{vaswani2017attention} is trained autoregressively to model the joint distribution over the text and image tokens. TediGAN \cite{xia2021tedigan} proposes to generate an image corresponding to a given text by training an encoder to map the text into the StyleGAN latent space. Several recent works \cite{crowson2022vqgan, abdal2022clip2stylegan} jointly utilize pre-trained generative models \cite{brock2018large, dhariwal2021diffusion, esser2021taming} and CLIP to steer the generated result towards the desired target description. More recently, diffusion models (DM) \cite{song2020improved, ho2020denoising, nichol2021improved} achieves state-of-the-art results on test-to-image synthesis by decomposing the image formation process into a sequential application of denoising autoencoders. Previous DMs operate directly in pixel space, often consuming hundreds of GPU training days. Latent diffusion models \cite{rombach2022high} are then proposed to enable DM training on limited computational resources while retaining their quality and flexibility by applying them to latent space. Later, \cite{nichol2021glide, liu2021more, avrahami2022blended} fill in the blank that previous DMs are mainly used to create abstract artworks from text descriptions and cannot edit parts of an actual image while preserving the rest.

\begin{figure*}[t]
    \centering
    \includegraphics[width=\linewidth]
    {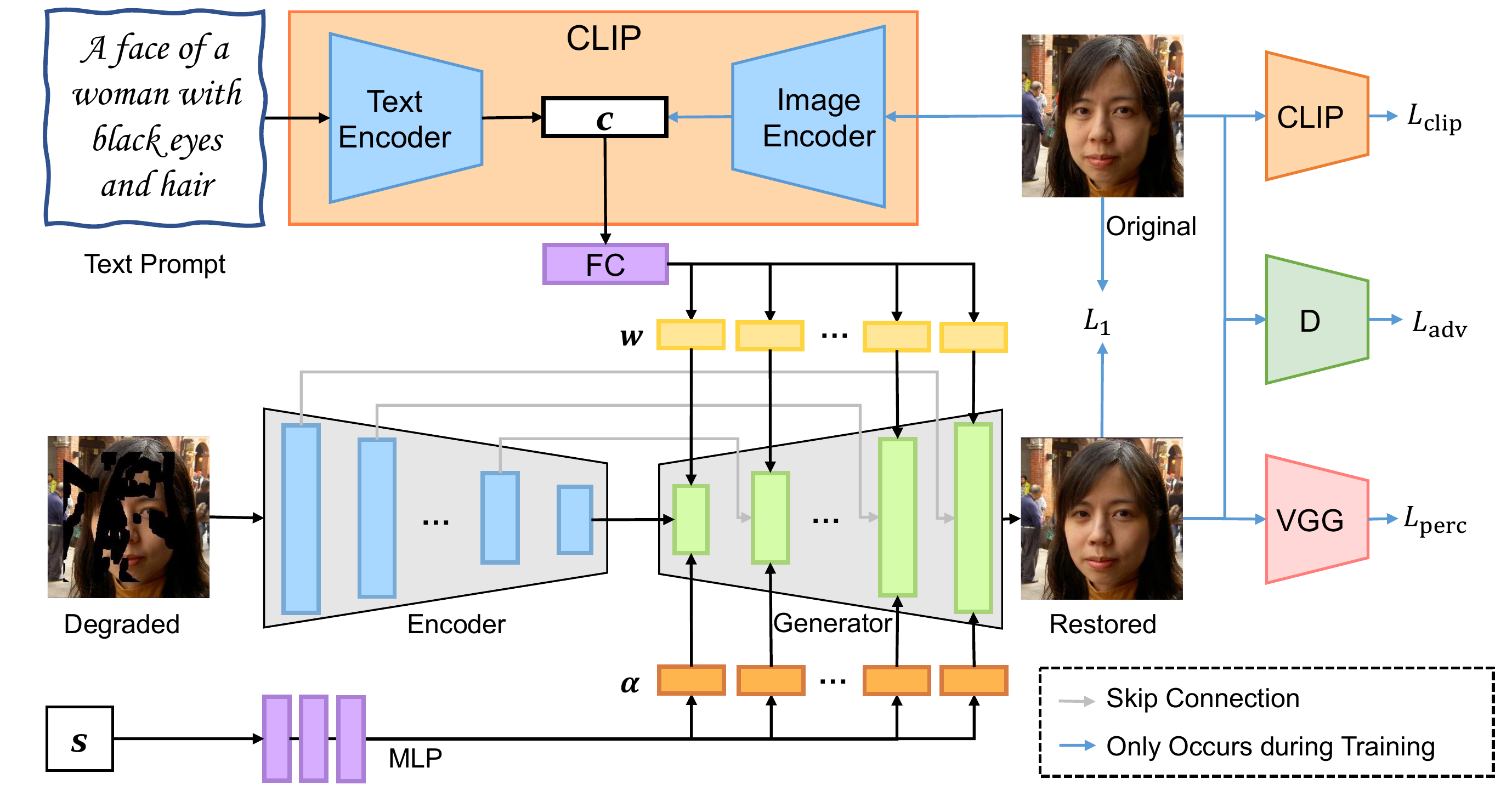}
    \caption{ \textbf{The proposed TextIR framework.} Our framework consists of an encoder and a generator. The encoder is used to extract features of the degraded images for fusion with the generated features. The generator is used to generate the restoration results. During training, we use the ground truth as the condition. With the help of CLIP's text-image shared feature space, we can use text as a condition to obtain results that match the description during inference.}
    \label{fig:arch1}   
\end{figure*}

%

\section{Proposed Method}
\label{sec:method}

Restoring a degraded image (e.g., masked image, grayscale image, low-resolution image) is an ill-posed problem because the missing information is uncertain. Our goal is to train an efficient model allowing the user to provide this information in text format and obtain a restored image corresponding to the text description. Text provides a highly intuitive user interface for describing the desired result. The most straightforward way to obtain such a model is to use text as conditional input during training and supervise the restored results with the corresponding image data. However, obtaining such text-image paired data is typically costly. Since CLIP's image and text feature spaces are semantically aligned, we can use the image features extracted by CLIP instead of text features as conditional inputs during training. Then, the conditional input image can be used as ground truth for supervision.






\subsection{Overall Pipeline}
TextIR takes as input the degraded image $I_d\in \mathbb{R}^{H_{in}\times W_{in}\times C_{in}}$ and a text condition $c$ that controls the attributes of the result, and outputs the restored image $I_r\in \mathbb{R}^{H_{out}\times W_{out}\times C_{out}}$ that satisfies the condition: $I_r = G(I_d, c)$. 
In the training process, we utilize the text-image shared space of CLIP to simulate texts with images, where the input of an image is similar to the input of the text. In this way, we can translate $I_{gt}$ into the most suitable ``text'' condition by CLIP's image encoder. The embeddings encoded by the CLIP encoder are taken as conditional input $c$ to the TextIR.
After training, the input condition can be converted into text embeddings.
The overall pipeline can be formalized as:
\begin{equation}
\begin{split}
&\text{training: }\ I_r = G(I_d, E_I(I_{gt})), \\
&\text{inference: }\ I_r = G(I_d, E_T(\text{text})),
\end{split}
\end{equation}
where $E_I$ and $E_T$ denote the image and text encoder of CLIP, respectively.

\subsection{Network Architecture}
TextIR consists of an encoder and a style-based generator. The encoder network is a simple convolution neural network (CNN) that takes the degraded image $I_d$ as input and extracts its multi-scale features $\{f^{0},...,f^{l-1}, f^{l}\}$. The feature map $f^l$ from the last layer is used as the ``constant'' input of style-based generator architecture. Encoded features from other layers are passed to the generator to be fused with the generated features of the same shape through skip connections. This practice ensures that the generated results match the input degraded images. The expression intensities of encoded and generated features are adjusted by a feature fusion module at each level under a conditional strength factor $s$.
The StyleConv layers \cite{karras2020analyzing} used in the generator also receive style code $\boldsymbol{w}$ for the modulate operation.
Since our model is a conditional generator, the style code $\boldsymbol{w}$ is obtained from the input condition $c$. 


\textbf{Style modulation.} StyleGANv2 \cite{karras2020analyzing} proposes style-modulated convolution to eliminate the drop-like artifacts in StyleGAN \cite{karras2019style}.  
The adaptive instance normalization (AdaIN) on feature maps is replaced by a weaker demodulation technique based on statistical assumptions of the input signal. It uses a style code to modulate the convolution kernel on the input channel dimension. Before conducting the convolution, the kernel is normalized channel-wisely to approximately preserve the standard deviation between the input and output features. We denote such a convolution operator as $\text{StyleConv}$. 

We apply $\text{StyleConv}$ to inject the conditions into the network.
The style code $\boldsymbol{w}=\left[w^{0},...,w_{1,2}^{l-1}, w_{1,2}^{l}\right]$ is obtained from the input text (inference) or image (training) embeddings $c$ by a fully-connected (FC) layer:
\begin{equation}
    w = \text{Reshape}(\text{FC}(c)),\ c = E_I(I_{gt})\ \ \text{or}\ \ c = E_T(\text{text}).
\end{equation}


Unlike StyleGANv2, our generator does not start from a constant value, but takes the feature $f^l$ from the encoder as input. Moreover, to fully utilize the information of the input image $I_d$, we fuse the multi-level features $\{f^{0},...,f^{l-1}, f^{l}\}$ output by encoder with their corresponding same shape generated features $\{g^{0},...,g^{l-1}, g^{l}\}$ by the feature fusion module to obtain $x^i$ at level $i$.
Then, the generator can generate the appropriate restoration part according to the degraded input.
Formally, the style modulation is defined as:
\begin{equation}
\small
    g^i =
    \left\{
        \begin{array}{ll}
        \text{StyleConv}(f^{l-i}, w^i) & i = 0, \\
        \text{StyleConv}(\text { StyleConv }(\uparrow_2(x^{i-1}), w^i_1), w^i_2) & i > 0,
        \end{array}
    \right.
\end{equation}
where $\uparrow_2$ denotes $2\times$ upsampling.

\begin{figure*}[t]
    \centering
    \includegraphics[width=\linewidth]
    {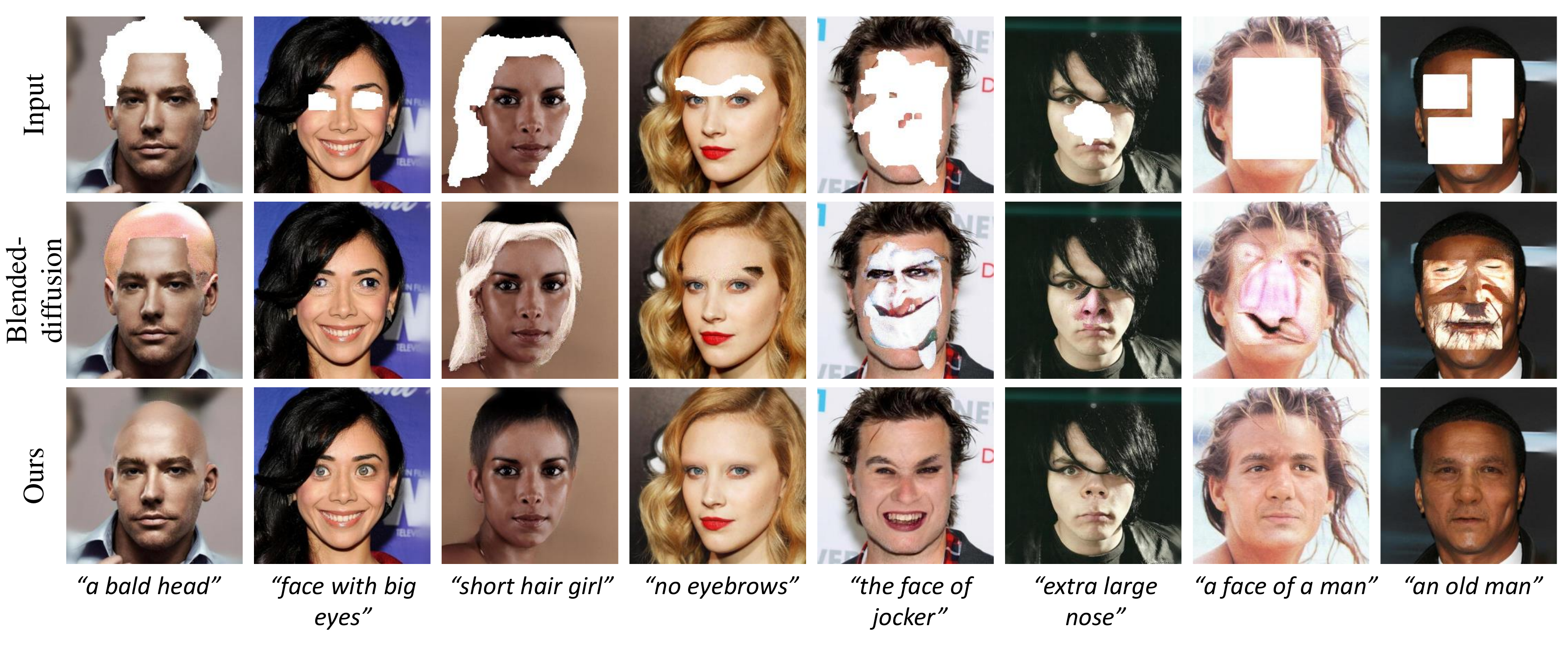}
    \caption{ \textbf{The inpainting results.} Compared to blended-diffusion, our method can produce a more natural and realistic result for masked images.}
    \label{fig:inpainting}   
\end{figure*}

\textbf{Feature fusion.} As mentioned before, we consider incorporating multi-level features from the encoder into the generator with skip connections. Since $I_d$ differ in the degree of degradation and the contribution of their features to the generation process, strength factors $s$ are employed to flexibly control the expression intensity of encoded and generated features \cite{he2022gcfsr}. Therefore, we use a channel-wise weighted sum for the feature fusion instead of simply concatenating or adding them together. Strength factors $s$ are first converted into a series of channel-wise weighting vectors by an MLP:
\begin{equation}
    \boldsymbol{\alpha} = \{\alpha_{1,2}^0,...,\alpha_{1,2}^{l-1},\alpha_{1,2}^{l}\} = \text{Reshape}(\text{MLP}(s)).
\end{equation}
At the $i$-th level, $\alpha_1^i$ and $\alpha_2^i\in\mathbb{R}^{chan(i)}$ are used as weights to fuse $f^{l-i}$ and $g^i$, where $chan(i)$ indicates the channel dimension. 
To avoid undesirable effects in the statistics of the output features of \text{StyleConv}, these weighting vectors are normalized to be positive and to have channel-wise unit $L_2$ norm. It is defined formally as:
\begin{equation}
\begin{split}
    &\alpha_{enc/gen}^i = \frac{|\alpha_{1/2}^i|}{\sqrt{{\alpha_1^i}^2 + {\alpha_2^i}^2 + \epsilon}},\ \ \alpha_{1/2}^i\in\boldsymbol{\alpha}, \\
    &x^i = \alpha_{enc}^i\cdot\text{Conv}(f^{l-i}) + \alpha_{gen}^i\cdot g^i,
\end{split}
\end{equation}
where $\epsilon$ equals to $10^{-8}$ and $\text{Conv}(\cdot)$ denotes the convolution operator for the initial adjustment of encoded features. In the super-resolution task, we set the strength factor $s$ as the downscaling factor. In other experiments, we also use the encoder to obtain $s$ by another MLP.


\subsection{Objective Functions}
We demonstrate the capability of our framework on three image restoration tasks: (\textbf{a}) image inpainting, (\textbf{b}) super-resolution, and (\textbf{c}) colorization. For training, we add the corresponding degradation on the ground-truth image $I_{gt}$ to get the degraded image $I_d$. In the case of inpainting, a mask $M$ is sampled from a free-form mask dataset \cite{liu2018image} to get $I_d=[I_{gt}\odot M, M]\in \mathbb{R}^{4\times 256\times 256}$, where $[\cdot, \cdot]$ denotes concatenation in the channel dimension and $\odot$ denotes Hadamard product. In the case of super-resolution, we downsample $I_{gt}$ and then upsample it to the original resolution to get $I_d\in \mathbb{R}^{3\times 512 \times 512}$. The downsampling factor $s$ is randomly sampled from $\{4, 8, 16, 32, 64\}$. In the case of colorization, we use the \textit{L} channel of the $I_{gt}$ in \textit{Lab} color space (composed of \textit{L}, \textit{a}, and \textit{b} channels) as $I_d\in \mathbb{R}^{1\times 256 \times 256}$, where the output of the network $I_o$ is the value of \textit{ab} channels. The output is concatenated together with $I_d$ and then converted to \textit{RGB} color space to get $I_r = lab\_to\_rgb([I_d, I_o])$.

We train all tasks using non-saturating adversarial loss:
\begin{equation}
\begin{split}
    \mathcal{L}_{adv,D} &= \mathbb{E}[\log(1 + \exp(-D(I_{gt}))) \\
    &+ \log(1 + \exp(D(G(I_d, c))))], \\
    \mathcal{L}_{adv,G} &= \mathbb{E}[\log(1+ \exp(-D(G(I_d, c))))],
\end{split}
\end{equation}
and introduce the CLIP loss to guide the restored result to satisfy the condition. We define $\mathcal{L}_{clip}$ as $1$ minus the cosine similarity of the result with $I_{gt}$ in the CLIP embedding space:
\begin{equation}
    \mathcal{L}_{clip} = 1 - \frac{E_I(I_r)\cdot E_I(I_{gt})}{|E_I(I_r)||E_I(I_{gt})|}.
\end{equation}

In combination with the $L_1$ loss (smooth $L_1$ in colorization) and the perceptual loss \cite{johnson2016perceptual}, the total loss is:
\begin{equation}
\small
\begin{split}
    &\mathcal{L}_D = \lambda_{adv}\mathcal{L}_{adv,D}, \\
    &\mathcal{L}_G = \lambda_{adv}\mathcal{L}_{adv,G} + \lambda_{clip}\mathcal{L}_{clip} + \lambda_{l_1}\mathcal{L}_1 + \lambda_{perc}\mathcal{L}_{perc}.
\end{split}
\end{equation}

We set $\lambda_{adv}=\lambda_{perc}=0.01,\lambda_{l_1}=1$ in all experiments, for inpainting, we set $\lambda_{clip}=0.5$, for super-resolution and colorization, we set $\lambda_{clip}=0.1$. For CLIP, we use the ViT-B/32 model to extract text prompt embeddings.

\begin{figure*}[t]
    \centering
    \includegraphics[width=\linewidth]
    {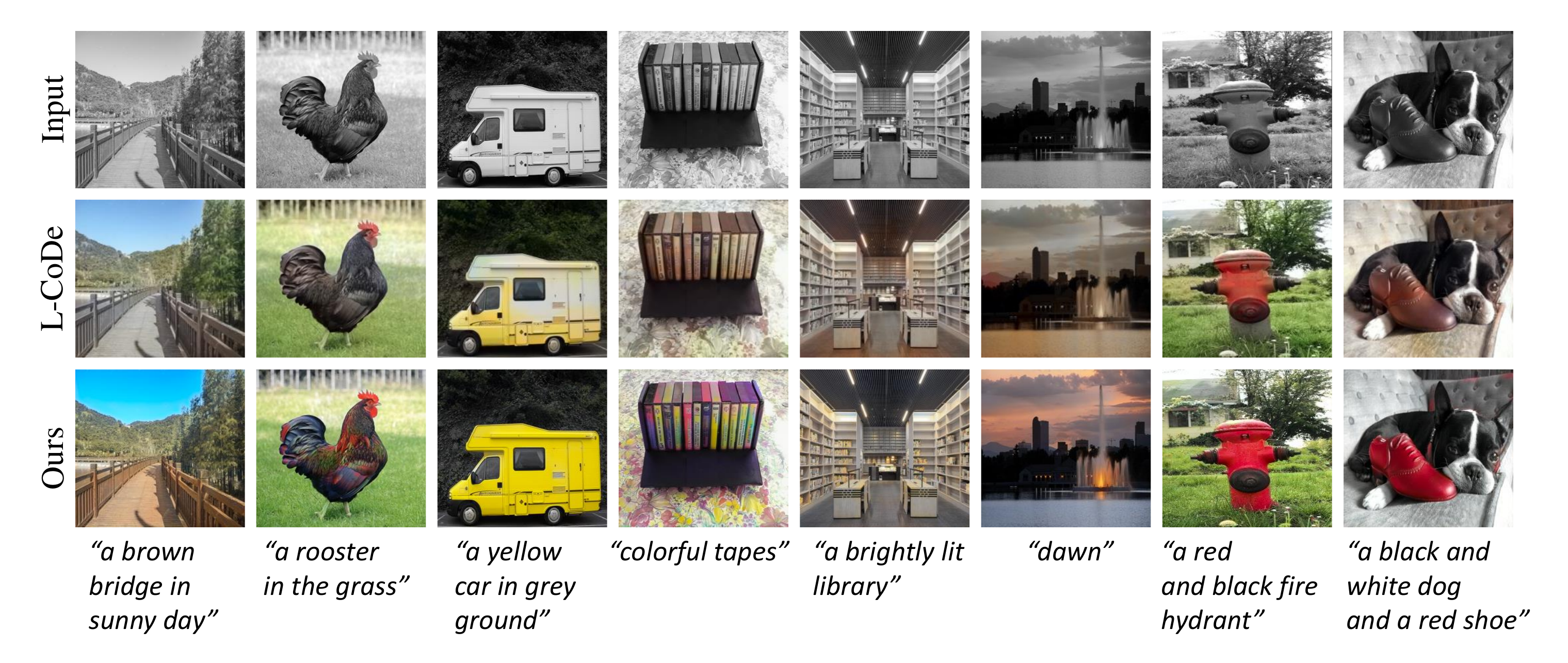}
    \caption{ \textbf{The colorization results.} Compared to L-CoDe, our method is able to locate the target to be colored more precisely. Our results are colorful and match the text descriptions.}
    \label{fig:colorization}   
\end{figure*}


\begin{figure*}[t]
    \centering
    \includegraphics[width=\linewidth]
    {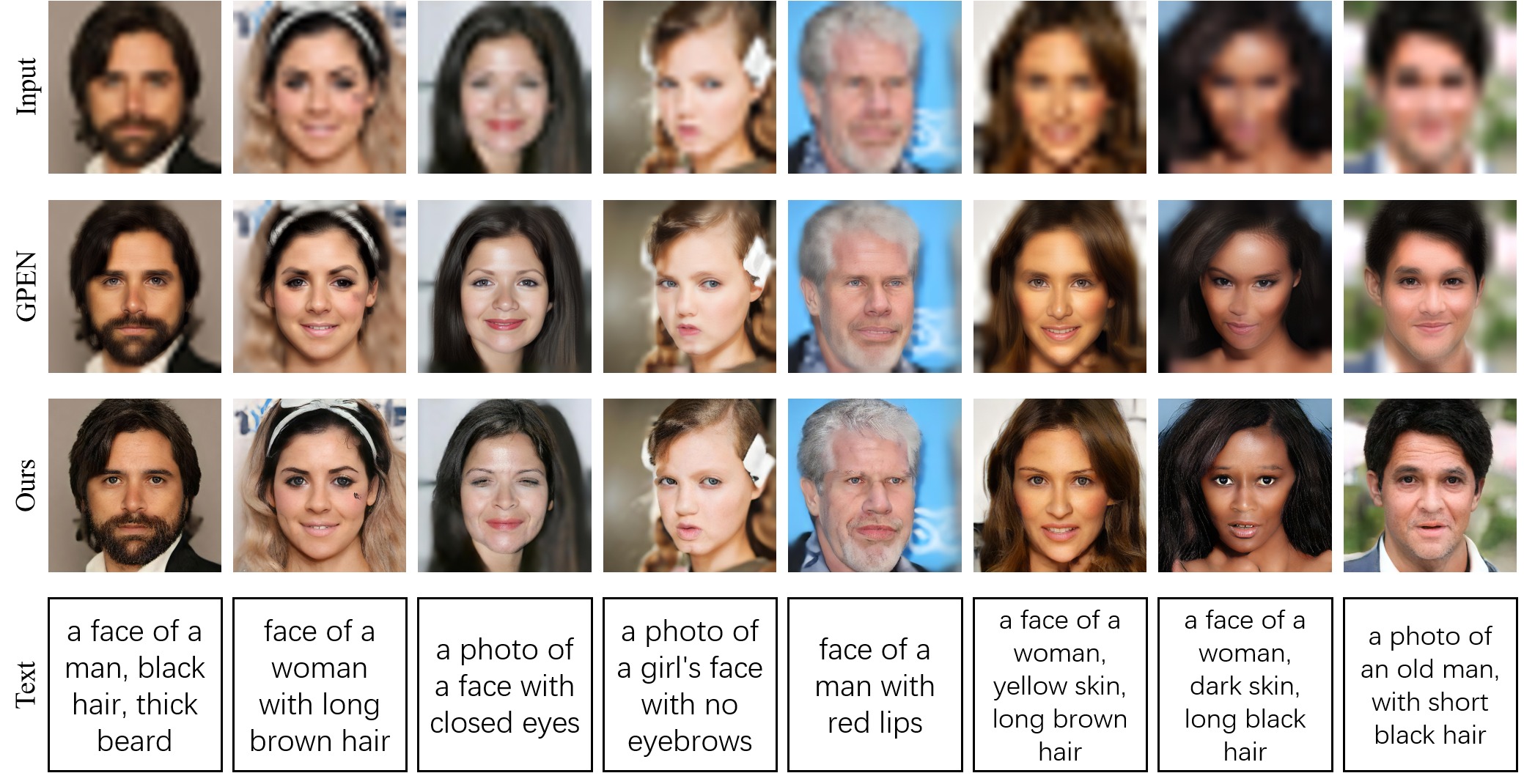}
    \caption{ \textbf{The super-resolution results.} Compared to GEPN, our method is able to recover more details and these details match the text description.}
    \label{fig:SR}   
\end{figure*}


\section{Experiments and Analysis}


\subsection{Implementation details}
We implement our model using the PyTorch framework. The optimizer we use is Adam \cite{kingma2014adam} and the learning rate for both encoder and generator is $2\times10^{-3}$. All models are trained on 2 NVIDIA Tesla V100 GPUs with mini-batch size of 16 for 300k iterations. Image inpainting and image super-resolution 
 experiments are performed on the FFHQ \cite{karras2019style} dataset with resolutions of $256\times256$ and $512\times512$, respectively. We use the ImageNet \cite{deng2009imagenet} with resolution of $256\times256$ for image colorization experiment. Image inpainting and super-resolution models are evaluated on the Multi-Modal-CelebA-HQ \cite{xia2021tedigan} dataset. Image colorization model is evaluated on COCO-Stuff \cite{caesar2018coco}. For evaluation, we adopt the widely used pixel-wise metrics (PSNR and SSIM) and the perceptual metric (LPIPS\cite{DBLP:conf/cvpr/ZhangIESW18}).


\begin{figure*}[t]
    \centering
    \includegraphics[width=\linewidth]
    {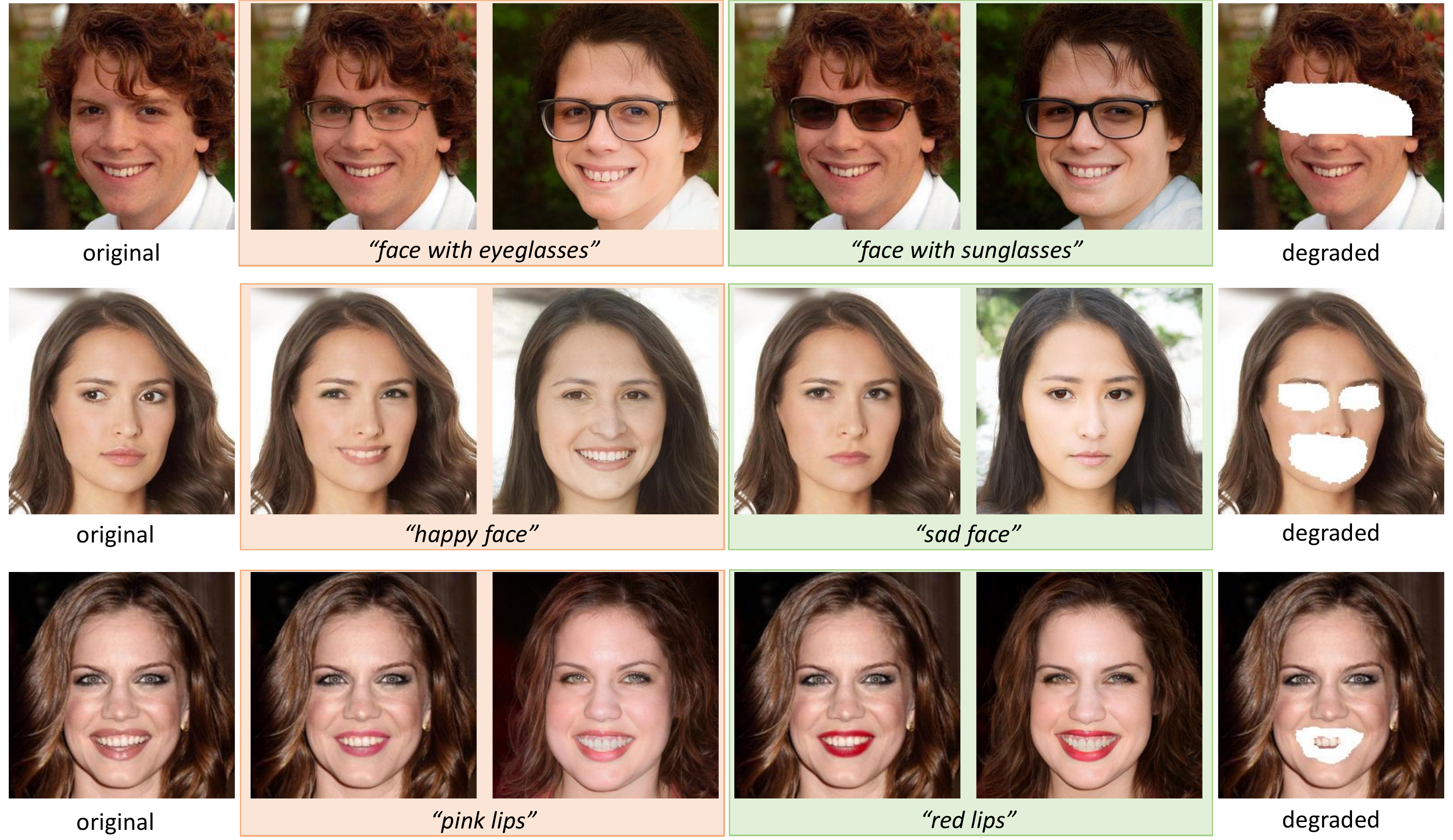}
    \caption{\textbf{Comparison with StyleCLIP on face editing}. We first mask out selected regions of the original face and then obtain the edited result by text-based image inpainting. Compared with GAN inversion-based methods like StyleCLIP, our method, \textbf{left side} of each rectangle, can specify local editing regions, thus keeping other regions unchanged and maintaining identity perfectly in face editing.}
    \label{fig:edit}   
\end{figure*}

\begin{figure*}[t]
    \centering
    \includegraphics[width=\linewidth]
    {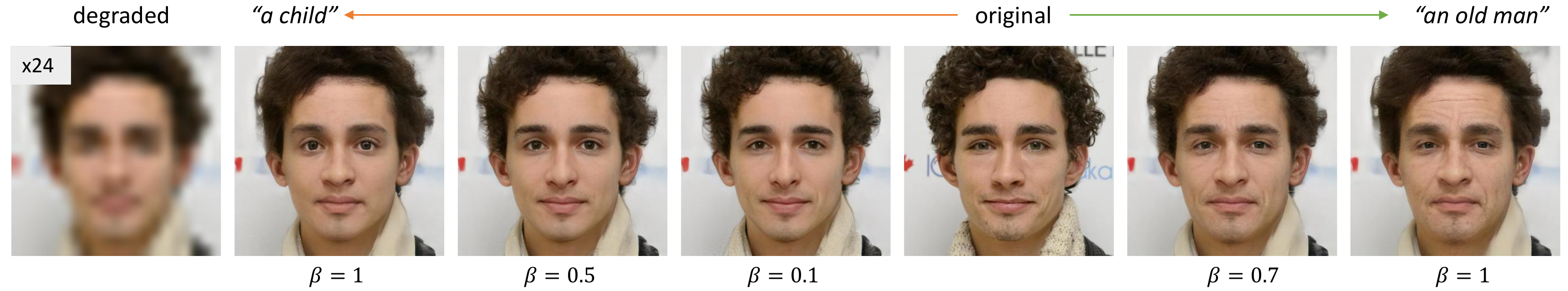}
    \caption{\textbf{Conditional strength interpolation.} We do interpolation between the CLIP embeddings of the edited text and the original image to get the control conditions of different intensities.}
    \label{fig:interp}   
\end{figure*}

\subsection{Image Inpainting}

In the experiments on image inpainting, there is no direct text-based image inpainting work available for comparison. Blended-diffusion \cite{avrahami2022blended} is a similar work. They use a denoising diffusion probabilistic model (DDPM) to generate natural-looking results and use CLIP to steer the edit results toward a target text description. They also use a mask to remove the content of the original image and replace it with the edited effect, but they have to use the original image as input. By replacing their image input with a masked image, their method can somehow have the capability of text-based inpainting. Then, we compare our method with theirs in this way.

We first qualitatively compare our method with the blended-diffusion. The faces with different regions masked out and a paired text input are fed into different models. The images restored according to the text are shown in Figure \ref{fig:inpainting}. The result of blended-diffusion looks less natural. For example, when inpainting the description of a bald head, the area inpainted by blended-diffusion is different from the original human skin color. When inpainting a man's face and the masked area is large, they do not make any meaningful content. In contrast, our results are more realistic and natural than their results, and all match the description of the text.


We quantitatively evaluate the two methods using the Multi-Modal-CelebA-HQ dataset, in which each image has several corresponding text labels. Using this label as input for the text condition, then the original image can be used as ground truth to evaluate the results. Test mask sampled from the irregular mask dataset \cite{liu2018image}. We use the average of these text labels for each image as condition input. The comparison results are shown in Table \ref{tab:quantitative1}. It can be seen that our method is much better in terms of these metrics compared to blended-diffusion.

\begin{table}
    \small
    \centering 
    \caption{Quantitative evaluation of inpainting experiment.}
\begin{tabular}{c|c|c|c} 
\hline
\multirow{2}{*}{ Methods } & \multicolumn{3}{c}{ Metrics } \\
\cline { 2 - 4 } & PSNR $\uparrow$ & SSIM$\uparrow$ & LPIPS$\downarrow$ \\
\hline
Blended-diffusion\cite{avrahami2022blended} & \makecell[c]{23.16}& \makecell[c]{0.901}& \makecell[c]{0.226} \\
Ours  & \makecell[c]{\textbf{29.83}}& \makecell[c]{\textbf{0.932}}& \makecell[c]{\textbf{0.068}} \\
\hline
\end{tabular}
    \label{tab:quantitative1}
\end{table}


\subsection{Image Colorization}

In the experiments of colorization, we compare our method with a previous language-based colorization method, L-CoDe\cite{weng2022code}. We also make a qualitative comparison first, and the comparison results are shown in Figure \ref{fig:colorization}. L-CoDe always fails to match the color to the target object. In the case of colorizing the car yellow, most of the car's body is still gray. When colorful colors were given to the tapes, their results were only grayish without distinct colors. In the last column, when the shoes were colorized with red, the red color was not obvious in their results. In contrast, our method colorizes the target image according to the text and accurately locates the target object. In the quantitative evaluation, we use the caption in COCO-Stuff as the text input and the corresponding image as the ground truth. The comparison results are shown in Table \ref{tab:quantitative2}. Our method exceed L-CoDe in all three metrics.



\begin{table}
    \small
    \centering 
    \caption{Quantitative evaluation of colorization experiment.}
\begin{tabular}{c|c|c|c} 
\hline
\multirow{2}{*}{ Methods } & \multicolumn{3}{c}{ Metrics } \\
\cline { 2 - 4 } & PSNR $\uparrow$ & SSIM$\uparrow$ & LPIPS$\downarrow$ \\
\hline
L-CoDe\cite{weng2022code} & \makecell[c]{24.965}& \makecell[c]{0.916}& \makecell[c]{0.169} \\
Ours  & \makecell[c]{\textbf{26.70}}& \makecell[c]{\textbf{0.923}}& \makecell[c]{\textbf{0.124}} \\
\hline
\end{tabular}
    \label{tab:quantitative2}
\end{table}

\subsection{ Image Super-resolution}


\begin{figure*}[h!]
    \centering
    \includegraphics[width=\linewidth]{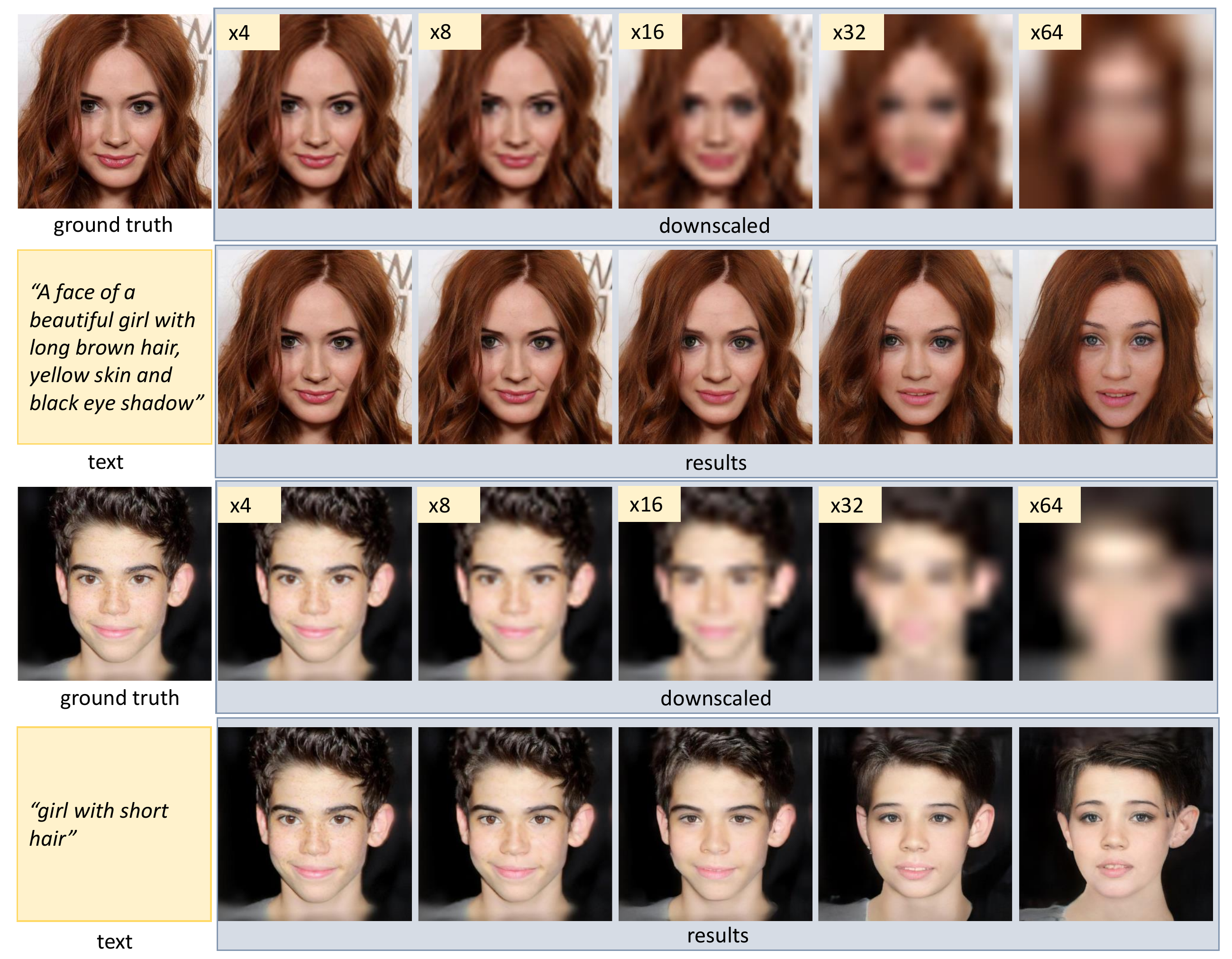}
    \caption{ \textbf{The ablation study of ``s''.} }
    \label{fig:sr_supp}   
\end{figure*}


For the experiment on image super-resolution, since there is no similar text-based method for comparison, we compare our methods with a blind face restoration method GPEN \cite{yang2021gan}. The results of the restored images are shown in Figure \ref{fig:SR}. Compared to the GPEN, the results of our method are clearer. The details are also consistent with the text descriptions. We also perform a quantitative comparison using the Multi-Modal-CelebA-HQ dataset in the same manner as above. The qualitative results for $16\times$ SR task are shown in Table \ref{tab:quantitative3}. The introduction of text information in the super-resolution process shows a significant improvement. 



\begin{table}
    \small
    \centering 
    \caption{Quantitative evaluation of super-resolution experiment.}
\begin{tabular}{c|c|c|c} 
\hline
\multirow{2}{*}{ Methods } & \multicolumn{3}{c}{ Metrics } \\
\cline { 2 - 4 } & PSNR $\uparrow$ & SSIM$\uparrow$ & LPIPS$\downarrow$ \\
\hline
GPEN\cite{yang2021gan} & \makecell[c]{26.82}& \makecell[c]{0.704}& \makecell[c]{0.273} \\
Ours  & \makecell[c]{\textbf{27.41}}& \makecell[c]{\textbf{0.784}}& \makecell[c]{\textbf{0.227}} \\
\hline
\end{tabular}
    \label{tab:quantitative3}
\end{table}

\subsection{Image Editing via Degradation}
TextIR is a text-based controllable restoration framework, so that we can perform text-based image editing by degrading first and restoring later. In contrast to traditional image editing paradigms, our degradation-based approach allows users to specify the area and degree of image degradation to preserve the specific information we want to keep. We compare our method with a previous text-based image editing method StyleCLIP\cite{patashnik2021styleclip}. Figure \ref{fig:edit} shows the face editing comparison of our approach with StyleCLIP's global direction method. Since our method only change the local area to be edited, it can perfectly preserve the identity information. While StyleCLIP can make edits based on text, it also changes other attributes and the identity.

\subsection{Ablation Studies}

\textbf{Interpolation.} By interpolating between the CLIP emdeddings of the original image and the text we want to edit, we can get a condition that is controlled in intensity with $\beta$: $c = \beta \cdot E_T(\text{text}) + (1-\beta)\cdot E_I(I_{gt})$. Figure \ref{fig:interp} shows an example, the results are obtained by image super-resolution. We can achieve fine-grained image manipulation by this way.

\textbf{Usage of ``s'' \& the help of text prior.} 
``s'' is used in all experiments to control the amount of the degraded image in the result according to the degradation level. As shown in Figure \ref{fig:sr_supp}. When the degradation is not severe, texts do not affect the image's identity. When the image degradation gets more severe, the role of text prior comes to the fore. 




\section{Conclusion}

In this work, we manage to use text information to assist in image restoration because text input is more readily available and provides information with higher flexibility. To achieve a text-based image restoration method, we utilize the recent CLIP model and design a simple and effective framework, which allows the user to use text input to get desired image restoration results. The framework utilize CLIP's text-image feature compatibility to enable a better fusion of image and text features. Our framework can be used for various image restoration tasks, including image inpainting, image super-resolution, and image colorization. 





{\small
\bibliographystyle{ieee_fullname}
\bibliography{egbib}

\begin{thebibliography}{10}\itemsep=-1pt

\bibitem{abdal2022clip2stylegan}
Rameen Abdal, Peihao Zhu, John Femiani, Niloy Mitra, and Peter Wonka.
\newblock Clip2stylegan: Unsupervised extraction of stylegan edit directions.
\newblock In {\em ACM SIGGRAPH 2022 Conference Proceedings}, pages 1--9, 2022.

\bibitem{abuolaim2020defocus}
Abdullah Abuolaim and Michael~S Brown.
\newblock Defocus deblurring using dual-pixel data.
\newblock In {\em European Conference on Computer Vision}, pages 111--126.
  Springer, 2020.

\bibitem{avrahami2022blended}
Omri Avrahami, Dani Lischinski, and Ohad Fried.
\newblock Blended diffusion for text-driven editing of natural images.
\newblock In {\em Proceedings of the IEEE/CVF Conference on Computer Vision and
  Pattern Recognition}, pages 18208--18218, 2022.

\bibitem{brock2018large}
Andrew Brock, Jeff Donahue, and Karen Simonyan.
\newblock Large scale gan training for high fidelity natural image synthesis.
\newblock {\em arXiv preprint arXiv:1809.11096}, 2018.

\bibitem{buhler2020deepsee}
Marcel~C Buhler, Andr{\'e}s Romero, and Radu Timofte.
\newblock Deepsee: Deep disentangled semantic explorative extreme
  super-resolution.
\newblock In {\em Proceedings of the Asian Conference on Computer Vision},
  2020.

\bibitem{caesar2018coco}
Holger Caesar, Jasper Uijlings, and Vittorio Ferrari.
\newblock Coco-stuff: Thing and stuff classes in context.
\newblock In {\em Proceedings of the IEEE conference on computer vision and
  pattern recognition}, pages 1209--1218, 2018.

\bibitem{chen2018language}
Jianbo Chen, Yelong Shen, Jianfeng Gao, Jingjing Liu, and Xiaodong Liu.
\newblock Language-based image editing with recurrent attentive models.
\newblock In {\em Proceedings of the IEEE Conference on Computer Vision and
  Pattern Recognition}, pages 8721--8729, 2018.

\bibitem{chen2018fsrnet}
Yu Chen, Ying Tai, Xiaoming Liu, Chunhua Shen, and Jian Yang.
\newblock Fsrnet: End-to-end learning face super-resolution with facial priors.
\newblock In {\em Proceedings of the IEEE Conference on Computer Vision and
  Pattern Recognition}, pages 2492--2501, 2018.

\bibitem{cho2021rethinking}
Sung-Jin Cho, Seo-Won Ji, Jun-Pyo Hong, Seung-Won Jung, and Sung-Jea Ko.
\newblock Rethinking coarse-to-fine approach in single image deblurring.
\newblock In {\em Proceedings of the IEEE/CVF international conference on
  computer vision}, pages 4641--4650, 2021.

\bibitem{crowson2022vqgan}
Katherine Crowson, Stella Biderman, Daniel Kornis, Dashiell Stander, Eric
  Hallahan, Louis Castricato, and Edward Raff.
\newblock Vqgan-clip: Open domain image generation and editing with natural
  language guidance.
\newblock In {\em European Conference on Computer Vision}, pages 88--105.
  Springer, 2022.

\bibitem{deng2009imagenet}
Jia Deng, Wei Dong, Richard Socher, Li-Jia Li, Kai Li, and Li Fei-Fei.
\newblock Imagenet: A large-scale hierarchical image database.
\newblock In {\em 2009 IEEE conference on computer vision and pattern
  recognition}, pages 248--255. Ieee, 2009.

\bibitem{dhariwal2021diffusion}
Prafulla Dhariwal and Alexander Nichol.
\newblock Diffusion models beat gans on image synthesis.
\newblock {\em Advances in Neural Information Processing Systems},
  34:8780--8794, 2021.

\bibitem{dudhane2022burst}
Akshay Dudhane, Syed~Waqas Zamir, Salman Khan, Fahad~Shahbaz Khan, and
  Ming-Hsuan Yang.
\newblock Burst image restoration and enhancement.
\newblock In {\em Proceedings of the IEEE/CVF Conference on Computer Vision and
  Pattern Recognition}, pages 5759--5768, 2022.

\bibitem{esser2021taming}
Patrick Esser, Robin Rombach, and Bjorn Ommer.
\newblock Taming transformers for high-resolution image synthesis.
\newblock In {\em Proceedings of the IEEE/CVF conference on computer vision and
  pattern recognition}, pages 12873--12883, 2021.

\bibitem{gu2019self}
Shuhang Gu, Yawei Li, Luc~Van Gool, and Radu Timofte.
\newblock Self-guided network for fast image denoising.
\newblock In {\em Proceedings of the IEEE/CVF International Conference on
  Computer Vision}, pages 2511--2520, 2019.

\bibitem{he2022gcfsr}
Jingwen He, Wu Shi, Kai Chen, Lean Fu, and Chao Dong.
\newblock Gcfsr: a generative and controllable face super resolution method
  without facial and gan priors.
\newblock In {\em Proceedings of the IEEE/CVF Conference on Computer Vision and
  Pattern Recognition}, pages 1889--1898, 2022.

\bibitem{ho2020denoising}
Jonathan Ho, Ajay Jain, and Pieter Abbeel.
\newblock Denoising diffusion probabilistic models.
\newblock {\em Advances in Neural Information Processing Systems},
  33:6840--6851, 2020.

\bibitem{jiang2021talk}
Yuming Jiang, Ziqi Huang, Xingang Pan, Chen~Change Loy, and Ziwei Liu.
\newblock Talk-to-edit: Fine-grained facial editing via dialog.
\newblock In {\em Proceedings of the IEEE/CVF International Conference on
  Computer Vision}, pages 13799--13808, 2021.

\bibitem{johnson2016perceptual}
Justin Johnson, Alexandre Alahi, and Li Fei-Fei.
\newblock Perceptual losses for real-time style transfer and super-resolution.
\newblock In {\em European conference on computer vision}, pages 694--711.
  Springer, 2016.

\bibitem{karras2019style}
Tero Karras, Samuli Laine, and Timo Aila.
\newblock A style-based generator architecture for generative adversarial
  networks.
\newblock In {\em Proceedings of the IEEE/CVF conference on computer vision and
  pattern recognition}, pages 4401--4410, 2019.

\bibitem{karras2020analyzing}
Tero Karras, Samuli Laine, Miika Aittala, Janne Hellsten, Jaakko Lehtinen, and
  Timo Aila.
\newblock Analyzing and improving the image quality of stylegan.
\newblock In {\em Proceedings of the IEEE/CVF conference on computer vision and
  pattern recognition}, pages 8110--8119, 2020.

\bibitem{kingma2014adam}
Diederik~P Kingma and Jimmy Ba.
\newblock Adam: A method for stochastic optimization.
\newblock {\em arXiv preprint arXiv:1412.6980}, 2014.

\bibitem{kwon2022clipstyler}
Gihyun Kwon and Jong~Chul Ye.
\newblock Clipstyler: Image style transfer with a single text condition.
\newblock In {\em Proceedings of the IEEE/CVF Conference on Computer Vision and
  Pattern Recognition}, pages 18062--18071, 2022.

\bibitem{li2018recurrent}
Xia Li, Jianlong Wu, Zhouchen Lin, Hong Liu, and Hongbin Zha.
\newblock Recurrent squeeze-and-excitation context aggregation net for single
  image deraining.
\newblock In {\em Proceedings of the European conference on computer vision
  (ECCV)}, pages 254--269, 2018.

\bibitem{liang2021swinir}
Jingyun Liang, Jiezhang Cao, Guolei Sun, Kai Zhang, Luc Van~Gool, and Radu
  Timofte.
\newblock Swinir: Image restoration using swin transformer.
\newblock In {\em Proceedings of the IEEE/CVF International Conference on
  Computer Vision}, pages 1833--1844, 2021.

\bibitem{lin2020mmfl}
Qing Lin, Bo Yan, Jichun Li, and Weimin Tan.
\newblock Mmfl: Multimodal fusion learning for text-guided image inpainting.
\newblock In {\em Proceedings of the 28th ACM International Conference on
  Multimedia}, pages 1094--1102, 2020.

\bibitem{liu2018image}
Guilin Liu, Fitsum~A Reda, Kevin~J Shih, Ting-Chun Wang, Andrew Tao, and Bryan
  Catanzaro.
\newblock Image inpainting for irregular holes using partial convolutions.
\newblock In {\em Proceedings of the European conference on computer vision
  (ECCV)}, pages 85--100, 2018.

\bibitem{liu2021more}
Xihui Liu, Dong~Huk Park, Samaneh Azadi, Gong Zhang, Arman Chopikyan, Yuxiao
  Hu, Humphrey Shi, Anna Rohrbach, and Trevor Darrell.
\newblock More control for free! image synthesis with semantic diffusion
  guidance.
\newblock {\em arXiv preprint arXiv:2112.05744}, 2021.

\bibitem{liu2019dual}
Xing Liu, Masanori Suganuma, Zhun Sun, and Takayuki Okatani.
\newblock Dual residual networks leveraging the potential of paired operations
  for image restoration.
\newblock In {\em Proceedings of the IEEE/CVF Conference on Computer Vision and
  Pattern Recognition}, pages 7007--7016, 2019.

\bibitem{lu2019vilbert}
Jiasen Lu, Dhruv Batra, Devi Parikh, and Stefan Lee.
\newblock Vilbert: Pretraining task-agnostic visiolinguistic representations
  for vision-and-language tasks.
\newblock {\em Advances in neural information processing systems}, 32, 2019.

\bibitem{mansimov2015generating}
Elman Mansimov, Emilio Parisotto, Jimmy~Lei Ba, and Ruslan Salakhutdinov.
\newblock Generating images from captions with attention.
\newblock {\em arXiv preprint arXiv:1511.02793}, 2015.

\bibitem{nichol2021glide}
Alex Nichol, Prafulla Dhariwal, Aditya Ramesh, Pranav Shyam, Pamela Mishkin,
  Bob McGrew, Ilya Sutskever, and Mark Chen.
\newblock Glide: Towards photorealistic image generation and editing with
  text-guided diffusion models.
\newblock {\em arXiv preprint arXiv:2112.10741}, 2021.

\bibitem{nichol2021improved}
Alexander~Quinn Nichol and Prafulla Dhariwal.
\newblock Improved denoising diffusion probabilistic models.
\newblock In {\em International Conference on Machine Learning}, pages
  8162--8171. PMLR, 2021.

\bibitem{patashnik2021styleclip}
Or Patashnik, Zongze Wu, Eli Shechtman, Daniel Cohen-Or, and Dani Lischinski.
\newblock Styleclip: Text-driven manipulation of stylegan imagery.
\newblock In {\em Proceedings of the IEEE/CVF International Conference on
  Computer Vision}, pages 2085--2094, 2021.

\bibitem{radford2021learning}
Alec Radford, Jong~Wook Kim, Chris Hallacy, Aditya Ramesh, Gabriel Goh,
  Sandhini Agarwal, Girish Sastry, Amanda Askell, Pamela Mishkin, Jack Clark,
  et~al.
\newblock Learning transferable visual models from natural language
  supervision.
\newblock In {\em International Conference on Machine Learning}, pages
  8748--8763. PMLR, 2021.

\bibitem{ramesh2022hierarchical}
Aditya Ramesh, Prafulla Dhariwal, Alex Nichol, Casey Chu, and Mark Chen.
\newblock Hierarchical text-conditional image generation with clip latents.
\newblock {\em arXiv preprint arXiv:2204.06125}, 2022.

\bibitem{ramesh2021zero}
Aditya Ramesh, Mikhail Pavlov, Gabriel Goh, Scott Gray, Chelsea Voss, Alec
  Radford, Mark Chen, and Ilya Sutskever.
\newblock Zero-shot text-to-image generation.
\newblock In {\em International Conference on Machine Learning}, pages
  8821--8831. PMLR, 2021.

\bibitem{razavi2019generating}
Ali Razavi, Aaron Van~den Oord, and Oriol Vinyals.
\newblock Generating diverse high-fidelity images with vq-vae-2.
\newblock {\em Advances in neural information processing systems}, 32, 2019.

\bibitem{reed2016generative}
Scott Reed, Zeynep Akata, Xinchen Yan, Lajanugen Logeswaran, Bernt Schiele, and
  Honglak Lee.
\newblock Generative adversarial text to image synthesis.
\newblock In {\em International conference on machine learning}, pages
  1060--1069. PMLR, 2016.

\bibitem{rombach2022high}
Robin Rombach, Andreas Blattmann, Dominik Lorenz, Patrick Esser, and Bj{\"o}rn
  Ommer.
\newblock High-resolution image synthesis with latent diffusion models.
\newblock In {\em Proceedings of the IEEE/CVF Conference on Computer Vision and
  Pattern Recognition}, pages 10684--10695, 2022.

\bibitem{song2020improved}
Yang Song and Stefano Ermon.
\newblock Improved techniques for training score-based generative models.
\newblock {\em Advances in neural information processing systems},
  33:12438--12448, 2020.

\bibitem{su2019vl}
Weijie Su, Xizhou Zhu, Yue Cao, Bin Li, Lewei Lu, Furu Wei, and Jifeng Dai.
\newblock Vl-bert: Pre-training of generic visual-linguistic representations.
\newblock {\em arXiv preprint arXiv:1908.08530}, 2019.

\bibitem{tan2019lxmert}
Hao Tan and Mohit Bansal.
\newblock Lxmert: Learning cross-modality encoder representations from
  transformers.
\newblock {\em arXiv preprint arXiv:1908.07490}, 2019.

\bibitem{van2017neural}
Aaron Van Den~Oord, Oriol Vinyals, et~al.
\newblock Neural discrete representation learning.
\newblock {\em Advances in neural information processing systems}, 30, 2017.

\bibitem{vaswani2017attention}
Ashish Vaswani, Noam Shazeer, Niki Parmar, Jakob Uszkoreit, Llion Jones,
  Aidan~N Gomez, {\L}ukasz Kaiser, and Illia Polosukhin.
\newblock Attention is all you need.
\newblock {\em Advances in neural information processing systems}, 30, 2017.

\bibitem{wang2022clip}
Can Wang, Menglei Chai, Mingming He, Dongdong Chen, and Jing Liao.
\newblock Clip-nerf: Text-and-image driven manipulation of neural radiance
  fields.
\newblock In {\em Proceedings of the IEEE/CVF Conference on Computer Vision and
  Pattern Recognition}, pages 3835--3844, 2022.

\bibitem{wang2021towards}
Xintao Wang, Yu Li, Honglun Zhang, and Ying Shan.
\newblock Towards real-world blind face restoration with generative facial
  prior.
\newblock In {\em Proceedings of the IEEE/CVF Conference on Computer Vision and
  Pattern Recognition}, pages 9168--9178, 2021.

\bibitem{wang2022uformer}
Zhendong Wang, Xiaodong Cun, Jianmin Bao, Wengang Zhou, Jianzhuang Liu, and
  Houqiang Li.
\newblock Uformer: A general u-shaped transformer for image restoration.
\newblock In {\em Proceedings of the IEEE/CVF Conference on Computer Vision and
  Pattern Recognition}, pages 17683--17693, 2022.

\bibitem{wei2022hairclip}
Tianyi Wei, Dongdong Chen, Wenbo Zhou, Jing Liao, Zhentao Tan, Lu Yuan, Weiming
  Zhang, and Nenghai Yu.
\newblock Hairclip: Design your hair by text and reference image.
\newblock In {\em Proceedings of the IEEE/CVF Conference on Computer Vision and
  Pattern Recognition}, pages 18072--18081, 2022.

\bibitem{weng2022code}
Shuchen Weng, Hao Wu, Zheng Chang, Jiajun Tang, Si Li, and Boxin Shi.
\newblock L-code: Language-based colorization using color-object decoupled
  conditions.
\newblock 2022.

\bibitem{xia2021tedigan}
Weihao Xia, Yujiu Yang, Jing-Hao Xue, and Baoyuan Wu.
\newblock Tedigan: Text-guided diverse face image generation and manipulation.
\newblock In {\em Proceedings of the IEEE/CVF conference on computer vision and
  pattern recognition}, pages 2256--2265, 2021.

\bibitem{xu2018attngan}
Tao Xu, Pengchuan Zhang, Qiuyuan Huang, Han Zhang, Zhe Gan, Xiaolei Huang, and
  Xiaodong He.
\newblock Attngan: Fine-grained text to image generation with attentional
  generative adversarial networks.
\newblock In {\em Proceedings of the IEEE conference on computer vision and
  pattern recognition}, pages 1316--1324, 2018.

\bibitem{yang2021gan}
Tao Yang, Peiran Ren, Xuansong Xie, and Lei Zhang.
\newblock Gan prior embedded network for blind face restoration in the wild.
\newblock In {\em Proceedings of the IEEE/CVF Conference on Computer Vision and
  Pattern Recognition}, pages 672--681, 2021.

\bibitem{yuan2021florence}
Lu Yuan, Dongdong Chen, Yi-Ling Chen, Noel Codella, Xiyang Dai, Jianfeng Gao,
  Houdong Hu, Xuedong Huang, Boxin Li, Chunyuan Li, et~al.
\newblock Florence: A new foundation model for computer vision.
\newblock {\em arXiv preprint arXiv:2111.11432}, 2021.

\bibitem{zamir2022restormer}
Syed~Waqas Zamir, Aditya Arora, Salman Khan, Munawar Hayat, Fahad~Shahbaz Khan,
  and Ming-Hsuan Yang.
\newblock Restormer: Efficient transformer for high-resolution image
  restoration.
\newblock In {\em Proceedings of the IEEE/CVF Conference on Computer Vision and
  Pattern Recognition}, pages 5728--5739, 2022.

\bibitem{zamir2020learning}
Syed~Waqas Zamir, Aditya Arora, Salman Khan, Munawar Hayat, Fahad~Shahbaz Khan,
  Ming-Hsuan Yang, and Ling Shao.
\newblock Learning enriched features for real image restoration and
  enhancement.
\newblock In {\em European Conference on Computer Vision}, pages 492--511.
  Springer, 2020.

\bibitem{zhang2017stackgan}
Han Zhang, Tao Xu, Hongsheng Li, Shaoting Zhang, Xiaogang Wang, Xiaolei Huang,
  and Dimitris~N Metaxas.
\newblock Stackgan: Text to photo-realistic image synthesis with stacked
  generative adversarial networks.
\newblock In {\em Proceedings of the IEEE international conference on computer
  vision}, pages 5907--5915, 2017.

\bibitem{zhang2018stackgan++}
Han Zhang, Tao Xu, Hongsheng Li, Shaoting Zhang, Xiaogang Wang, Xiaolei Huang,
  and Dimitris~N Metaxas.
\newblock Stackgan++: Realistic image synthesis with stacked generative
  adversarial networks.
\newblock {\em IEEE transactions on pattern analysis and machine intelligence},
  41(8):1947--1962, 2018.

\bibitem{zhang2020text}
Lisai Zhang, Qingcai Chen, Baotian Hu, and Shuoran Jiang.
\newblock Text-guided neural image inpainting.
\newblock In {\em Proceedings of the 28th ACM International Conference on
  Multimedia}, pages 1302--1310, 2020.

\bibitem{DBLP:conf/cvpr/ZhangIESW18}
Richard Zhang, Phillip Isola, Alexei~A. Efros, Eli Shechtman, and Oliver Wang.
\newblock The unreasonable effectiveness of deep features as a perceptual
  metric.
\newblock In {\em {IEEE} Conference on Computer Vision and Pattern
  Recognition,{CVPR}}, pages 586--595. Computer Vision Foundation / {IEEE}
  Computer Society, 2018.

\bibitem{zhang2019residual}
Yulun Zhang, Kunpeng Li, Kai Li, Bineng Zhong, and Yun Fu.
\newblock Residual non-local attention networks for image restoration.
\newblock {\em arXiv preprint arXiv:1903.10082}, 2019.

\end{thebibliography}
}

\end{document}